\documentclass[conference]{IEEEtran}
\usepackage{fancyhdr}
\IEEEoverridecommandlockouts

\usepackage{algpseudocode}
\usepackage{algorithm}  
\usepackage{dblfloatfix}
\usepackage{setspace}
\usepackage{amsmath}
\allowdisplaybreaks[4]
\usepackage{graphicx}   
\usepackage{nomencl}
\usepackage{diagbox}
\usepackage{multirow}
\usepackage{subfigure}
\usepackage{bm} 
\usepackage{makecell}
\usepackage[utf8]{inputenc}
\usepackage[T1]{fontenc}
\usepackage{booktabs}
\usepackage{geometry}
\usepackage{amssymb}
\usepackage{cancel}
\usepackage{mathrsfs}
\def\BibTeX{{\rm B\kern-.05em{\sc i\kern-.025em b}\kern-.08em
    T\kern-.1667em\lower.7ex\hbox{E}\kern-.125emX}}
\begin{document}

%\title{Autonomous Edge-Cloud Cooperation for DNN Inference via Reinforcement Learning}
\title{Edge-Cloud Cooperation for DNN Inference via Reinforcement Learning and Supervised Learning}

% \author{Paper \#1570746810 by Anonymous Authors
% }
\author{\IEEEauthorblockN{Tinghao Zhang \IEEEauthorrefmark{1}, Zhijun Li \IEEEauthorrefmark{2}, Yongrui Chen \IEEEauthorrefmark{3}, Kwok-Yan Lam \IEEEauthorrefmark{1}, Jun Zhao \IEEEauthorrefmark{1}}
\IEEEauthorblockA{\IEEEauthorrefmark{1} School of Computer Science and Engineering, Nanyang Technological University, Singapore\\
                \{tinghao001, kwokyan.lam, junzhao\}@ntu.edu.sg \\
                \IEEEauthorrefmark{2} School of Computer Science and Technology, Harbin Institute of Technology, Harbin, China\\
                lizhijun\_os@hit.edu.cn\\
                \IEEEauthorrefmark{3} Department of Electronic and Communication Engineering, University of Chinese academy of sciences, Beijing, China\\
                chenyr@ucas.ac.cn\\
                }}
% \author{\IEEEauthorblockN{Tinghao Zhang\IEEEauthorrefmark{1},
%  Zhijun Li\IEEEauthorrefmark{2}, Yongrui Chen\IEEEauthorrefmark{3}, 
% Kwok-Yan Lam\IEEEauthorrefmark{4}, and Jun Zhao \IEEEauthorrefmark{1}}
% \IEEEauthorblockA{Department of Whatever,
% Whichever University\\
% Wherever\\
% Email: \IEEEauthorrefmark{1}author.one@add.on.net,
% \IEEEauthorrefmark{2}author.two@add.on.net,
% \IEEEauthorrefmark{3}author.three@add.on.net,
% \IEEEauthorrefmark{4}author.four@add.on.net}}
% \markboth{Journal of \LaTeX\ Class Files,~Vol.~14, No.~8, August~2015}%
% {Shell \MakeLowercase{\textit{et al.}}: Bare Demo of IEEEtran.cls for IEEE Journals}

\maketitle

\thispagestyle{fancy}
\pagestyle{fancy}
\lhead{This paper appears in the Proceedings of 2022 IEEE International Conferences on Internet of Things (iThings).\\ Please feel free to contact us for questions or remarks.} 

\cfoot{~\\[-25pt]\thepage}

\begin{abstract}
Deep Neural Networks (DNNs) have been widely applied in Internet of Things (IoT) systems for various tasks such as image classification and object detection. However, heavyweight DNN models can hardly be deployed on edge devices due to limited computational resources. In this paper, an edge-cloud cooperation framework is proposed to improve inference accuracy while maintaining low inference latency. To this end, we deploy a lightweight model on the edge and a heavyweight model on the cloud. A reinforcement learning (RL)-based DNN compression approach is used to generate the lightweight model suitable for the edge from the heavyweight model. Moreover, a supervised learning (SL)-based offloading strategy is applied to determine whether the sample should be processed on the edge or on the cloud. Our method is implemented on real hardware and tested on multiple datasets. The experimental results show that (1) The sizes of the lightweight models obtained by RL-based DNN compression are up to $87.6\%$ smaller than those obtained by the baseline method; (2) SL-based offloading strategy makes correct offloading decisions in most cases; (3) Our method reduces up to $78.8\%$ inference latency and achieves higher accuracy compared with the cloud-only strategy.
\end{abstract}

\begin{IEEEkeywords}
Edge-cloud Cooperation, Reinforcement Learning, Supervised Learning, Data Offloading.
\end{IEEEkeywords}

\IEEEpeerreviewmaketitle

\section{Introduction}~\label{intro}
In the 5G and beyond 5G era, future Internet of Things (IoT) systems evolve towards a smarter and more versatile paradigm. To this end, deep neural network (DNN) models have been widely applied in IoT systems to carry out sophisticated tasks such as computer vision, natural language processing, and optimal control. However, executing DNN models is computation expensive as it usually generates massive amounts of data. To deal with this high computational task, cloud computing has been widely used to conduct DNN inference. Specifically, data generated from end devices (e.g. smartphones or Raspberry Pis) will be uploaded to a centralized cloud, and then inference results will be sent back to the end devices. Although computation latency is alleviated with the help of cloud computing, high communication latency is introduced as tremendous amounts of data are constantly transmitted between devices and the cloud. Under this circumstance, edge computing is proposed to address this issue. In edge computing, data will be processed on edge devices instead of being sent to the cloud center. The latency is reduced significantly since computational resources are located close to the end devices. Therefore, edge computing plays a critical role in IoT systems.
    
Nowadays, simple and shallow models can hardly satisfy increasing users' demands. Thus, DNN models are designed to be much deeper and more heavyweight. For example, ResNet-18 and ResNet-50 contain about 11 million and 23 million trainable parameters, respectively. Edge devices are far from sufficient to support those cumbersome DNN models because of their limited computational capability, which implies that cloud computing cannot be overlooked. Thus, it is essential to develop an effective edge-cloud cooperation framework for IoT systems to tackle complex tasks. Intuitive wisdom for implementing edge-cloud cooperation is to deploy a lightweight DNN model and a heavyweight model on the edge and cloud, respectively. In this case, the edge devices will have enough resources to execute lightweight models, and the heavyweight model also participates in DNN inference. For obtaining lightweight models, we resort to DNN compression techniques. However, DNN compression is intractable due to the implicit connection between network architectures and the performance of the model. Thus, the hand-crafted design of lightweight models usually degrades their performance. In addition to DNN compression, an offloading strategy is also required to decide whether the input data should be processed on the edge or on the cloud.

%(我建议有针对性的写一段目前的边云协同架构的问题分析，从而引出下面的工作)
% More interestingly, previous study has ...有的数据适合边，有的数据适合云，这个方法正好可以决定哪个该云哪个改变→识别率比单边/云更高 introduction

In this article, we propose an edge-cloud cooperation framework that contains two key modules: reinforcement learning (RL)-based DNN compression and a supervised learning (SL)-based offloading strategy. First, RL is used to generate a lightweight model (called a student model) from a heavyweight model (called a teacher model), and the student model is trained using the knowledge distillation technique. The student model is deployed on the edge while the teacher model is deployed on the cloud. The computation capability of edge devices is adequate to execute the student model. As compressing DNN models usually harms the performance of the inference, the teacher model will provide correct inference for certain samples but the student model won't. These data should be offloaded to the cloud to ensure the overall accuracy of the systems. To this end, we formulate the data offloading problem between the edge and the cloud as a binary classification. A supervised learning model is used to determine which data should be offloaded to the cloud. The SL model is deployed on the edge devices together with the student model. We make sure that the SL model is lightweight and has little effect on inference speed. Besides, the experimental results of the previous work~\cite{Anubhav18} have shown that the student model can make correct predictions for certain samples while the teacher model cannot. Therefore, it's interestingly observed that the proposed edge-cloud cooperation framework achieves even higher overall accuracy than the original teacher model.

%  or on the cloud. Out method is implemented on real hardware and tested on multiple datasets. The experimental results show that (1) The sizes of the lightweight models obtained by RL-based DNN compression are up to $87.6\%$ smaller than those obtained by the baseline method; (2) SL-based offloading strategy makes correct offloading decisions in most cases; (3) Out method reduces up to $78.8\%$ inference latency and achieves higher accuracy compared with the cloud-only strategy.

The main contributions of this work are summarized as follows:
\begin{enumerate}
    \item We propose a RL-based DNN compression approach, where a reward function is designed to minimize compression ratios without sacrificing much accuracy. During training the RL agent, knowledge distillation is adopted to train the student model. The well-trained student model is deployed on the edge.

    \item We formulate data offloading as a binary classification problem and solve it by SL models. Specifically, SL models are used to predict whether the inference result of the student model is correct. If correct, the data are processed on the edge. Otherwise, the data are offloaded to the cloud. Therefore, autonomous edge-cloud cooperation is realized with the help of SL-based offloading strategies. More importantly, since the offloading strategies alleviate the deterioration of accuracy caused by DNN compression, the proposed DNN compression approach can attach more emphasis on compression ratio, thus further speeding up DNN inference on the edge.

    \item We show the effectiveness of the proposed edge-cloud cooperation framework on multiple datasets. Using the RL-based DNN compression method, the model size is up to $87.6\%$ smaller than that compressed by the baseline method. SL-based offloading strategies enable the IoT systems to achieve higher accuracy compared with the teacher model. Overall, the proposed edge-cloud cooperation framework reduces up to $78.8\%$ inference latency and achieves higher accuracy compared with the cloud-only strategy.

\end{enumerate}

The rest of the paper is organized as follows. System model is defined in Section~\ref{section3}. The details of the RL-based DNN compression approach and SL-based offloading strategies are introduced in Section~\ref{section4} and Section~\ref{sl}, respectively. Experimental results are provided in Section~\ref{section6}. The related work is reviewed in Section~\ref{section2} followed by the conclusion in Section~\ref{section7}.

\begin{figure}[t]
  \centering
  \includegraphics[width=8.5cm]{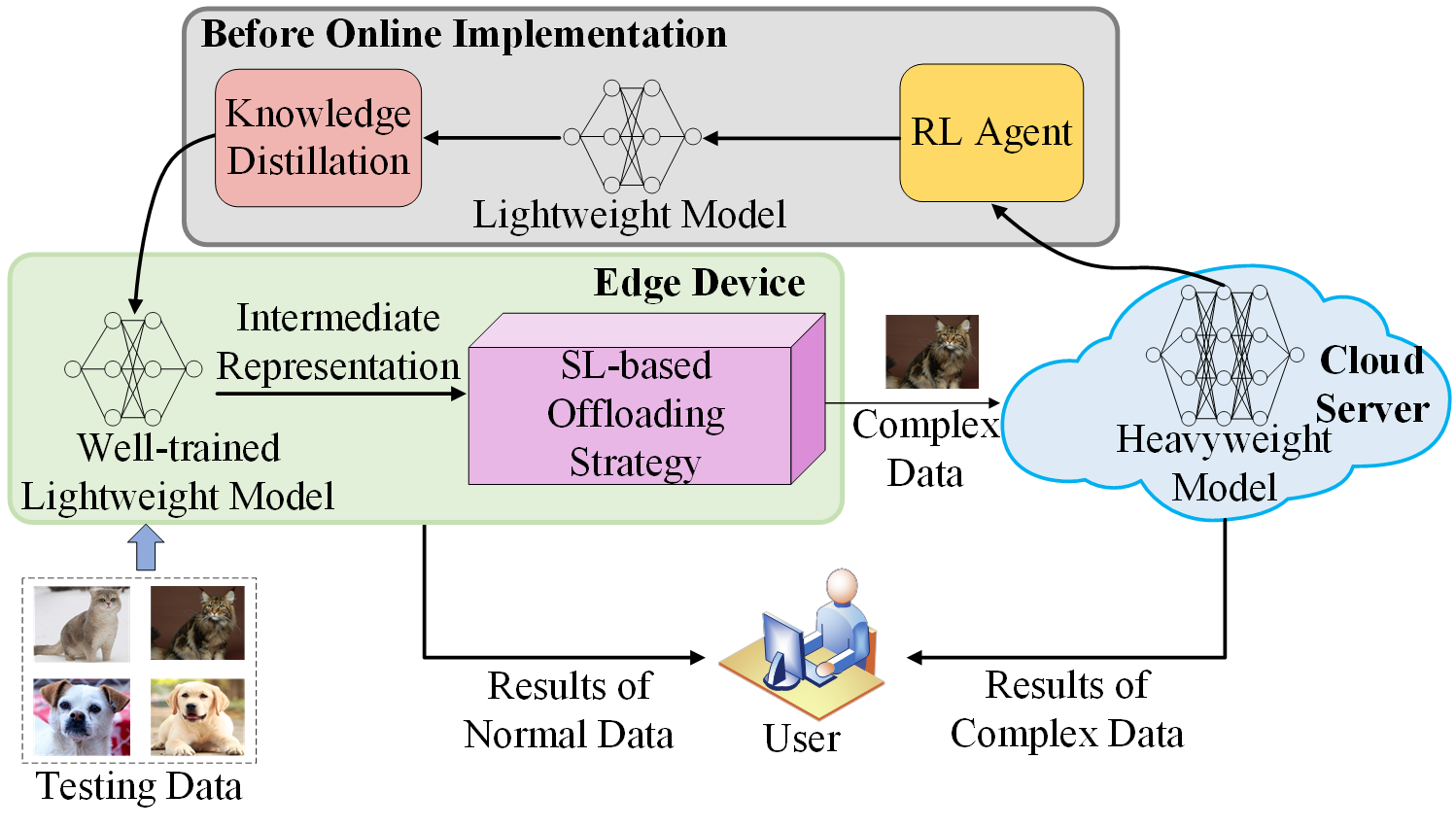}
\caption{Overview of the proposed Edge-cloud Cooperation System}
\label{overall}
\end{figure}

\section{Related Work}~\label{section2}
Our work falls into the classes of improvement methods for DNN inference and data offloading strategy.

DNN partition and DNN compression are two common approaches for the implementation of edge-cloud cooperation frameworks. In terms of DNN partition, a DNN model is partitioned into several parts which will be executed between edge and cloud separately. 
%DNN inference is expected to cost less time as data are processed fast in the cloud. 
The authors in \cite{Kang17} propose a lightweight scheduler named Neurosurgeon that can automatically partition a DNN model. In \cite{En18}, DNN models are trained by BranchyNet~\cite{Surat17} to achieve early exit, and a joint optimization algorithm is proposed to find the partition point and exit point. By leveraging BranchyNet, the author in~\cite{Teerapittayanon17} propose distributed deep neural networks and a joint training method to achieve low latency for edge-cloud systems. In~\cite{Zhang20}, DNN partition is formulated as a min-cut problem in a directed acyclic graph (DAG), and a two-stage approach is presented to solve the problem. However, DNN partition has to transmit high-dimensional intermediate features between the edge and cloud frequently, thus leading to unavoidable data transfer delay. Moreover, resource-limited edge devices also affect the performance of DNN partition. 

In this paper, we resort to DNN compression to essentially address this problem. There have been several recent studies that conduct DNN compress using various techniques. In~\cite{Song16}, the authors prune the unnecessary connections of the DNN models, and weights quantification and Huffman coding are employed to compress the model. In~\cite{Kim16}, the authors propose a one-shot whole network compression approach and implement it on the smartphone. In~\cite{Yang19}, an end-to-end DNN training framework is proposed to formulate the DNN training as an optimization problem, and an approximate algorithm is used to solve the problem. The authors in~\cite{Anubhav18} use reinforcement learning to remove and shrink the layers of DNN models. However, \cite{Anubhav18} requires two individual reinforcement learning agents, which complicates the training process. In addition, most existing DNN compression methods fail to bridge the gap between DNN compression and edge-cloud cooperation. 

We investigate offloading strategies to overcome the mentioned shortage of DNN compression. Traditional data offloading between the edge and the cloud usually refers to computation offloading and resource allocation~\cite{Liang19,Fariba20,Jiang20}. In the above studies, offloading strategies are proposed to minimize the execution latency given the resource constraint and amounts of data. In our work, the computation resources of the edge devices are sufficient enough for executing the student models, and the accuracy of the student model may be affected by DNN compression. In this circumstance, the offloading strategy focuses on improving the accuracy of the edge-cloud system. To achieve this, we employ supervised learning (SL) algorithms as the offloading strategies. To the best of our knowledge, it is an early effort to employ SL for offloading strategies based on the accuracy of the student models.

\section{System Model}~\label{section3}
In this section, we outline the proposed edge-cloud cooperation framework.

Take an image classification task as an example, an overview of the proposed edge-cloud cooperation framework is depicted in Fig.~\ref{overall}. The dataset is represented as $\mathcal{D} = \left\{(\bm{x}_i \in \mathbb{R}^d,y_i) \right\}^{\vert \mathcal{D} \vert}_{i=1}$ with $\vert \mathcal{D} \vert$ data samples, where $\bm{x}_i$ and $y_i$ denote the $i$-th data and its corresponding label, respectively. 
% （$y_i$是实数集？）

Before online implementation, RL-based DNN compression is carried out to obtain a student model. By using knowledge distillation, the well-trained student model is obtained. The student model along with a SL-based offloading strategy is deployed on the edge device. The teacher model is deployed on the cloud server. When the edge device receives a batch of samples (e.g., images), the student model is first used to determine the class labels. Note that the performance of the student model usually deteriorates when DNN compression achieves a high compression ratio. As a result, some data are too complex for the student model to make correct inference. We define this kind of data as \textbf{\emph{complex data}}. Complex data are represented as $\mathcal{D}_{c} = \left\{(\bm{x}_i \in \mathbb{R}^d,y_i) | 1 \leq i \leq \vert \mathcal{D} \vert, \hat{y}_i \neq y_i \right\}$, where $\hat{y}_i$ is the predicted label given by the student model. We define $\mathcal{D}_{N} = \complement_{\mathcal{D}}\mathcal{D}_{c}$ as normal data of the student model, where $\complement_{\mathcal{D}}$ represents a complement of a set with a universal set $\mathcal{D}$. The SL-based offloading strategy is used to decide which data belong to complex data. The data regarded by the SL-based offloading strategy as complex data are represented as $\mathcal{D}_{SL}$. Finally, the cloud server uses the teacher model to conduct DNN inference and returns results to the edge. In the following sections, more technical details, such as how to define the state information and how to train the SL-based offloading strategy, are provided.
% （DN的表示看不懂；图的results of normal data的箭头起始位置不清晰）
% 分流：一部分中间，一部分normal data
% Entire accuracy, 正确上云比 

\section{Reinforcement Learning-based DNN Compression}~\label{section4}
In this section, we introduce the background knowledge on RL and outline the details of the RL-based DNN compression method.

\subsection{Reinforcement Learning}
RL has made remarkable achievements in optimization control and decision making problems. In RL, the problems are usually described as Markov Decision Processes (MDPs). At each time step, a RL agent obtains state information $s_t$ from the environment. The RL then takes an action $a_t$ after analyzing current state information. A reward $r_t$ and the state information of next time step $s_{t+1}$ are obtained. The cumulative discounted reward $R$ of RL is derived as: 
\begin{equation}
  R=\Sigma^T_{t=0}{\gamma^{t}r(s_t,a_t)}
\end{equation}
where $T$ is the length of the trajectory. RL keeps updating the policy network $\pi_\theta$ in order to maximize expected cumulative discounted future reward $\mathbb{E}_{\pi_\theta}\left[R\right]$.

% \begin{figure}[t]
%   \centering
%   \includegraphics[width=8.5cm]{RL.png}
% \caption{Workflow of generating the student model using RL-based DNN Compression}
% \label{RL}
% \end{figure}

\subsection{RL-based DNN compression}

Layers of a DNN model can be shrunk or removed sequentially, which is a sequential decision making problem. Thus, DNN compression is modeled as a MDP, and RL is used to deal with this problem. To employ reinforcement learning to conduct DNN compression, four important elements should be defined: agent, state, action, and reward.

\begin{itemize}
    \item \textbf{Agent}: In this paper, gate recurrent unit (GRU) is employed as the RL agent to make sequential decision, since GRU has fewer parameters than long short-term memory (LSTM) while maintaining similar performance. By using GRU, the action of each step is determined by current state, previous state, and previous action.
    
    \item \textbf{State}: $s_t$ denotes the state information of the $t$-th layer and is defined as 
        \begin{equation}
          s_t = (l, k, d, p, o)
        \end{equation}
        where $l$ denotes the layer type, $k$ is kernel size, $d$ is stride, $p$ is padding, and $o$ means the number of outputs. For a neural network, each layer is represented by a 5-dimensional vector. Given a neural network with $T$ layers in total, there will be $T$ states in an episode (i.e., $s_0, ..., s_{T-1}$).
        
    \item \textbf{Action}: In this paper, we adopt a discrete action space to enable the RL agent to shrink or remove the layers of the DNN model. To this end, we design a candidate pool $\Phi$ containing several coefficients as follows: 
        \begin{equation}
              \Phi = \{ 0, 0.2, 0.4, 0.6, 0.8, 1\}
        \end{equation}
    The action $a_t$ is the index of the coefficients. For the $t$-th layer (i.e., $s_t$), the agent chooses a coefficient $\phi$ from the pool. The layer is removed if $\phi = 0$. Otherwise, the state of this layer is shrunk as follows:
    \begin{equation}
      s'_{t} = (l,k \cdot \phi,d \cdot \phi, p \cdot \phi, o \cdot \phi)
      \label{action}
    \end{equation}
    
    \item \textbf{Reward:} The reward function directly guides the training process of the RL agent. The reward is 0 for those intermediate states, and the discount factor $\gamma$ is 1, which means DNN compression has not been finished yet. After DNN compression, the reward will be calculated for the whole episode.
    
    In our work, DNN compression is expected to achieve a high compression ratio without sacrificing much accuracy. Therefore, we define the reward functions as shown in equation~\ref{reward}.
    \begin{equation} 
    C = 1 - \frac{\#\text{size}(\bm{w})}{\#\text{size}(\widehat{\bm{w}})}
    \label{reward}
    \end{equation}
    \begin{equation}
        R_C= e^{\alpha \cdot (C-C_0)},\ R_A= e^{\beta \cdot (A-A_0)}
    \end{equation}
    \begin{equation} 
    R = R_A \cdot R_C
    \label{reward}
    \end{equation}
   where $\widehat{\bm{w}}$ is the teacher model, $\bm{w}$ is the student model, $C$ is compression ratio, $A$ is the accuracy of the student model, $\alpha$ and $\beta$ are the constants, $C_0$ and $A_0$ are compression and accuracy threshold respectively, and $R_C$ and $R_A$ are compression and accuracy rewards, respectively.
   
   The reward function enables the RL agent to find a highly compressed student model that maintains acceptable performance. Moreover, there will be a heavy penalty when $C$ or $A$ are less than their corresponding thresholds. $\alpha$ and $\beta$ determine the weights between accuracy and compression. $C_0$ and $A_0$ are determined based on the task requirements, while $\alpha$, and $\beta$ are tuned empirically through our experiments.
\end{itemize}

\subsection{Training Process}
An effective optimization method is required to train the policy. Reference \cite{Anubhav18} has demonstrated that vanilla policy gradient (VPG), so called REINFORCE~\cite{Williams92}, outperform actor critic when the recurrent policy is used in DNN compression. Therefore, we adopt VPG to train the policy $\pi_\theta$ as follows. 
\begin{equation} 
\begin{array}{rcl}
\nabla_\theta J_{\theta}  & = & \nabla \mathbb{E}_{\tau \sim \pi_{\theta}(\tau)}\left[ R_\tau \right]\\
                & = & \mathbb{E}_{\tau \sim \pi_{\theta}(\tau)}\left[\nabla_\theta \log \pi_{\theta}(\tau) R_\tau\right]\\
                & = & \frac{1}{N}\sum\limits^N_{i=0}\sum\limits^K_{t=0}\left[\nabla_\theta \log \pi_{\theta}(a_{i,t}|s_{i,t})R_i \right]  \\
\end{array} \end{equation}
where $N$ is the batch size, $K$ denotes the number of layers. In order to reduce the variance, a baseline is introduced by
\begin{equation} 
\nabla_\theta J_{\theta}   =  \frac{1}{N}\sum\limits^N_{i=0}\sum\limits^K_{t=0}\left[\nabla_\theta \log \pi_{\theta}(a_{i,t}|s_{i,t})(R_i-b)  \right] 
\end{equation}
where b is an exponential moving average of the previous rewards. As a result, the policy can be updated by
\begin{equation}
  \theta \leftarrow \theta + \varepsilon\nabla_\theta J(\theta) 
  \label{updatepolicy}
\end{equation}
where $\varepsilon$ is the learning rate.

The reinforcement learning-based DNN compression approach is summarized in Algorithm~\ref{weidai}.

\begin{algorithm}[t]
\textsl{}\setstretch{1}
\caption{Reinforcement Learning-based DNN Compression}  
\begin{algorithmic}[1] 

\Require{Candidate pool $\Phi$, teacher model $\widehat{\bm{w}}$, Maximum training episodes $N$, the number of layers $T$, the RL agent $\pi_{\theta}$}

\Ensure Optimal student model $\bm{w}^\ast$
\State Initialize a reward list $\Omega = \{-\infty\}$
\For{ $i$ = 1 to $N$} 
    \State Initialize a student model $\bm{w}_{i} = \widehat{\bm{w}}$
    \For{ $t$ = 0 to $T-1$ }  
        \State Obtain $s_{t}$ based on the $t$-th layer of $\bm{w}_{i}$
        \State Generate action $a_t$ using $\pi_\theta(s_{t-1})$
        \State Select the coefficient $\phi$ from the $\Phi$ based on $a_t$
        \State Compress the $t$-th layer of $\bm{w}_{i}$ according to 
        \Statex \ \ \ \ \ \ \ \;\;Equation~\ref{action}
    \EndFor

    \State Train $\bm{w}_{i}$ by knowledge distillation
    \State Calculate $R_i$ according to Equation~\ref{reward}
    \If {$R_i > \text{max}\ \Omega$}
        \State $\bm{w}^\ast = \bm{w}_{i}$
    \EndIf
    \State $\Omega \leftarrow R_i$
    \State Update $\pi_\theta$ according to Equation~\ref{updatepolicy}
\EndFor

\State \Return $\bm{w}^\ast$

\end{algorithmic}\label{weidai}
\end{algorithm}  

\section{Supervised Learning-based Offloading Strategy}~\label{sl}
\subsection{Selecting SL model}
The offloading strategy is the key technology to determine which data should be offloaded to the cloud. The offloading strategy is deployed on the edge devices and is expected to predict whether a sample belongs to normal data or complex data. The data will be sent and processed on the cloud only if the offloading strategy classifies them as complex data. Based on the above discussion, the offloading problem can be formulated as a binary classification problem, and we adopt SL models as the offloading strategies to handle this classification problem.

\begin{figure}[t]
  \centering
  \includegraphics[width=8cm]{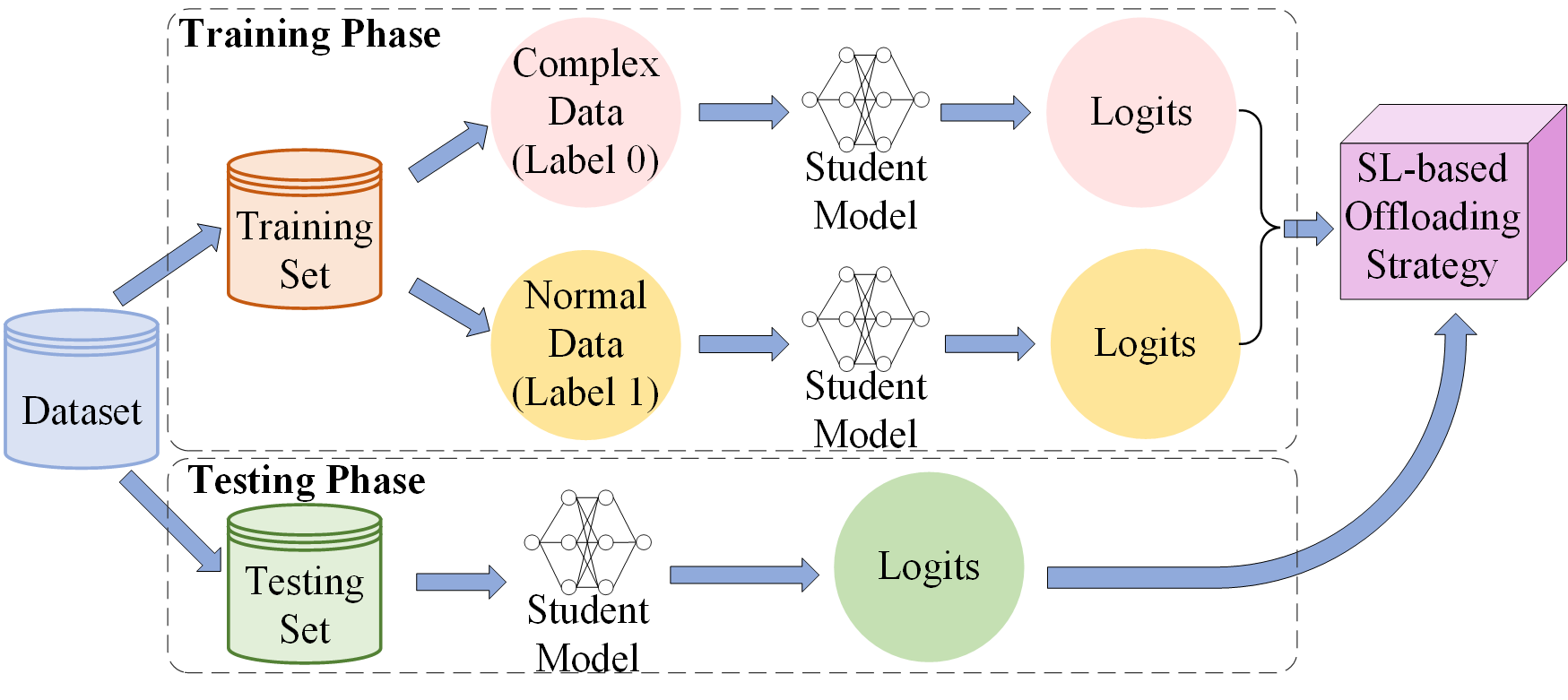}
\caption{Training and Testing Phases of the SL-based Offloading Strategy}
\label{offloading2}
\end{figure}

In addition to accuracy, the limited computation resources of the edge devices should be also taken into consideration when designing the offloading strategy. In other words, the size of the SL-based offloading strategy should be small enough to avoid high computation latency. Thus, we resort to simple but effective machine learning (ML) models such as support vector machine (SVM), random forest (RF), and k-nearest neighbors algorithm (KNN). To determine which SL models are the optimal offloading strategy, accuracy is the primary criterion as it directly affects the overall accuracy of the entire system. Besides, it is critical to balance the tradeoff between accuracy and the size of SL models to make sure that the offloading strategies will not significantly affect the execution time on the edge. Based on the experimental results in Section~\ref{section6}, the size of SVM is the smallest among other SL models in most cases, and SVM achieves sub-optimal accuracy on multiple datasets. Therefore, in this work, SVM is recommended as a general SL model for offloading strategies in real implementation.

\subsection{Training SL model with QBC-based data selection}
To begin with, labels are required to be provided before training SL models. In our work, the data will be labeled as 1 if the inference result of the student model is right. Otherwise, the data are complex data and their labels are 0. Intuitively, the number of complex data depends on the performance of the student model. Next, SL models are trained to predict the labels. As SL models have difficulty dealing with high-dimensional data like image data, the logits of the data are used as the feature vectors to train the SL models. As shown in Figure~\ref{offloading2}, the complex data and the normal data are selected from the original training dataset to train the SL models. Finally, the SL models are tested by the testing set. 

A salient point to note is that the number of complex data in the training set is less than that of normal data for SL models. For example, given a student model with an accuracy of 80\%, 20\% of the data are complex data, and normal data account for 80\% of the data. Since imbalanced training data affect model performance, we use the same number of normal data to train SL models. Therefore, the approach to select normal data affects the performance of SL models. Instead of completely random selection, we adopt the query by committee (QBC) algorithm~\cite{Seung92} to select those normal data which are the most informative for the models. In the QBC algorithm, a score $S(x)$ is assigned for a sample $x$ with a class $c \in C$ ($C$ denotes a set of possible classes) by a committee that includes $L$ classifiers, and the sample with the highest score will be selected. In this work, the entropy $e_{l}(x)$ is used to assign the score $S(x)$ via equation~\ref{entropy} and~\ref{score}. 
\begin{equation}
  e_{l}(x) =-\sum\limits_{c}P_l(c|x)\log P_l(c|x), l = 1,2,...,L
  \label{entropy}
\end{equation}

\begin{equation}
  S(x) = \max(D_1(x),D_2(x),...,D_L(x))
  \label{score}
\end{equation}
where $P_l(c|x)$ denotes the probability of the class $c$ given the example $x$ and the $l$-th classifier, and $D_{l}(x)$ denotes the entropy of of the class distribution. 

\begin{algorithm}[t]
\textsl{}\setstretch{1}
\caption{Normal data selection using the QBC algorithm}  
\begin{algorithmic}[1]
\Require{Normal data $\mathcal{D}_{N}$, complex data $\mathcal{D}_{C}$, number of complex data $M$, a committee containing $L$ classifiers, $r$}

\Ensure Optimal normal data $\mathcal{D}^{*}_{N}$ for training the SL-based
\Statex \quad\ \;\; offloading strategy
\State $\mathcal{D}^{*}_{N} \leftarrow \{\}$ ;
\State Randomly select $r \cdot M$ samples from $\mathcal{D}_{N}$, add these samples to $\mathcal{D}^{*}_{N}$ and remove them from $\mathcal{D}_{N}$;

\State Train $L$ classifiers using $\mathcal{D}_{C}$ and $\mathcal{D}^{*}_{N}$;
\For{ each sample $x$ in $\mathcal{D}_{N}$}  
    \For{ $l$ = 1 to $L$}  
            \State Calculate the entropy $e_{l}(x)$ according to 
            \Statex \quad \quad \ \ \ \ \ \ \ Equation \ref{entropy};
    \EndFor
        
    \State Obtain the score $S(x)$ according to Equation~\ref{score};
        
\EndFor
    
    \State Sort the samples of $\mathcal{D}_{N}$ in descending order based
    \Statex \quad \ \  on their scores;
    \State Remove the first $(1-r) \cdot M$ samples from $\mathcal{D}_{N}$ and
    \Statex \quad \ \  add them to the $\mathcal{D}^{*}_{N}$

\State \Return $\mathcal{D}^{*}_{N}$
\end{algorithmic}\label{weidaima2}
\end{algorithm}

In our work, given the number of complex data $M$, $r \cdot M$ normal data are first selected randomly, where $0<r<1$. Next, the QBC algorithm is used to select $(1-r) \cdot M$ normal data. The details of the QBC-based sample selection approach are shown in Algorithm 2.

\begin{table*}[]
\centering
\caption{Performance Evaluation of DNN compression (ResNet-18 is the teacher model)}
\renewcommand\arraystretch{1}
\setlength{\tabcolsep}{4mm}{
\begin{tabular}{cccccc}\toprule
\multirow{2}{*}{Method}          & \multirow{2}{*}{Dataset} & \multicolumn{2}{c}{Teacher Model} & \multicolumn{2}{c}{Student Model} \\
                                 &                          & Size            & Accuracy        & Size            & Accuracy        \\\midrule
\multirow{3}{*}{Proposed Method} & MNIST                    & 43704 KB        & 99.33\%          & 407 KB          & 99.49\%          \\
                                 & SVHN                     & 43708 KB        & 95.22\%          & 491 KB          & 93.44\%          \\
                                 & CIFAR-10                 & 43707 KB        & 89.32\%          & 675 KB          & 84.12\%          \\
\multirow{2}{*}{N2N}             & SVHN                     & 43724 KB        & 92.01\%          & 3963 KB         & 91.81\%          \\
                                 & CIFAR-10                 & 44332 KB        & 95.24\%          & 2239 KB         & 95.38\%         \\\bottomrule 
\end{tabular}}
\label{tab1}
\end{table*}

\begin{table*}[]
\centering
\caption{Performance Evaluation of DNN compression (ResNet-34 is the teacher model)}
\renewcommand\arraystretch{1}
\setlength{\tabcolsep}{4mm}{
\begin{tabular}{cccccc}\toprule
\multirow{2}{*}{Method}          & \multirow{2}{*}{Dataset} & \multicolumn{2}{c}{Teacher Model} & \multicolumn{2}{c}{Student Model} \\
                                 &                          & Size            & Accuracy        & Size            & Accuracy        \\\midrule
\multirow{3}{*}{Proposed Method} & MNIST                    & 83236 KB        & 99.53\%          & 818 KB          & 99.44\%          \\
                                 & SVHN                     & 83240 KB        & 95.27\%          & 785 KB          & 92.91\%          \\
                                 & CIFAR-10                 & 83240 KB        & 92.59\%          & 1769 KB         & 86.19\%          \\
N2N                              & CIFAR-10                 & 83255 KB        & 92.35\%          & 8156 KB         & 92.35\%        \\\bottomrule  
\end{tabular}}
\label{tab2}
\end{table*}

\section{PERFORMANCE EVALUATION}~\label{section6}
\subsection{Experiment Setup}
The cloud equipment is an 8-core Intel core i7-9700k@3.60GHz processor and an NVIDIA RTX 2080i TX GPU. The edge device is the Raspberry Pi 4B platform as the edge end device, which is equipped with a 4-core BCM2835@1.5GHz processor and 4G RAM. For DNN compression, the policy network of the RL agent is a two-layer GRU model, and the learning rate is set as 0.0001. For QBC algorithms, $r = 0.5$, and $L = 3$. We adopt six candidate SL models as the offloading strategies: SVM, RF, KNN, gradient boosting decision tree (GBDT), extreme gradient boosting (XGB), and extra trees (ET).

\subsection{DNN Compression Performance}
The RL-based DNN compression approach is evaluated on a ResNet-18 model and on a ResNet-34 network. Three public datasets, MNIST, CIFAR-10, and the Street View House Numbers (SVHN) dataset, are used in our paper.
% \begin{itemize}
%     \item \textbf{MNIST}: MNIST~\cite{Cun90} consists of small square 28$\times$28 pixel gray-scale images of handwritten single digits between 0 and 9. 60,000 training images and 10,000 testing images are used for the experiments.
%     \item \textbf{CIFAR-10}: CIFAR-10~\cite{Krizhevsky09} contains 60,000 32$\times$32 color images covering ten object classes, and each class has 6,000 images. We use 50,000 training images and 10,000 testing images for the experiments. 
%     \item \textbf{SVHN}: The Street View House Numbers~\cite{Yuval11} dataset contains 32$\times$32 colored digit images with 10 classes. SVHN consists of 73257 digits for training, 26032 digits for testing.
% \end{itemize}

First, ResNet-18 and ResNet-34 are trained as the teacher models, respectively. Then, RL-based DNN compression is carried out using the teacher models. $\alpha$ is set as 20, $\beta$ is set as 15, and the batch size $N$ is set as 2 on CIFAR-10 and 3 on MNIST and SVHN. Next, similarity-preserving knowledge distillation is used to train the student model~\cite{Tung19}. Student models are trained for 80 epochs on CIFAR-10 and for 40 epochs on MNIST and SVHN. We compared the results with N2N learning in the table \ref{tab1} and \ref{tab2}. Note that the results of N2N learning are cited from its original paper~\cite{Anubhav18}. 

Unlike N2N learning, our work places compression ratio as the first priority when designing the reward function. This is because the offloading strategy can send complex data to the cloud center to ensure the accuracy of the entire system. Table \ref{tab1} and~\ref{tab2} show the details of the teacher models and the student models. We observe that all student models obtain a high compression ratio. On MNIST, our method achieves both high accuracy and high compression ratio. On SVHN for ResNet-18, the accuracy of N2N learning is 1.94\% higher than that of our method; the model size of our method is 87.6\% smaller than that of N2N learning. For CIFAR-10, the accuracy of N2N learning is 7.69\% and 6.16\% higher than that of our method regarding ResNet-18 and ResNet-34 as the teacher models, respectively; the model sizes of our method are 69.9\% and 78.3\% smaller than those of N2N learning regarding ResNet-18 and ResNet-34 as the teacher models, respectively. Compared with N2N learning, our method successfully obtains a higher compression ratio while slightly reducing models' accuracy.

\subsection{Offloading Strategy Performance}
\begin{figure*}[t]
\centering
\subfigure[MNIST (ResNet-18 is the teacher model)]{
\begin{minipage}[t]{0.33\linewidth}
\centering
\includegraphics[width=5.6cm]{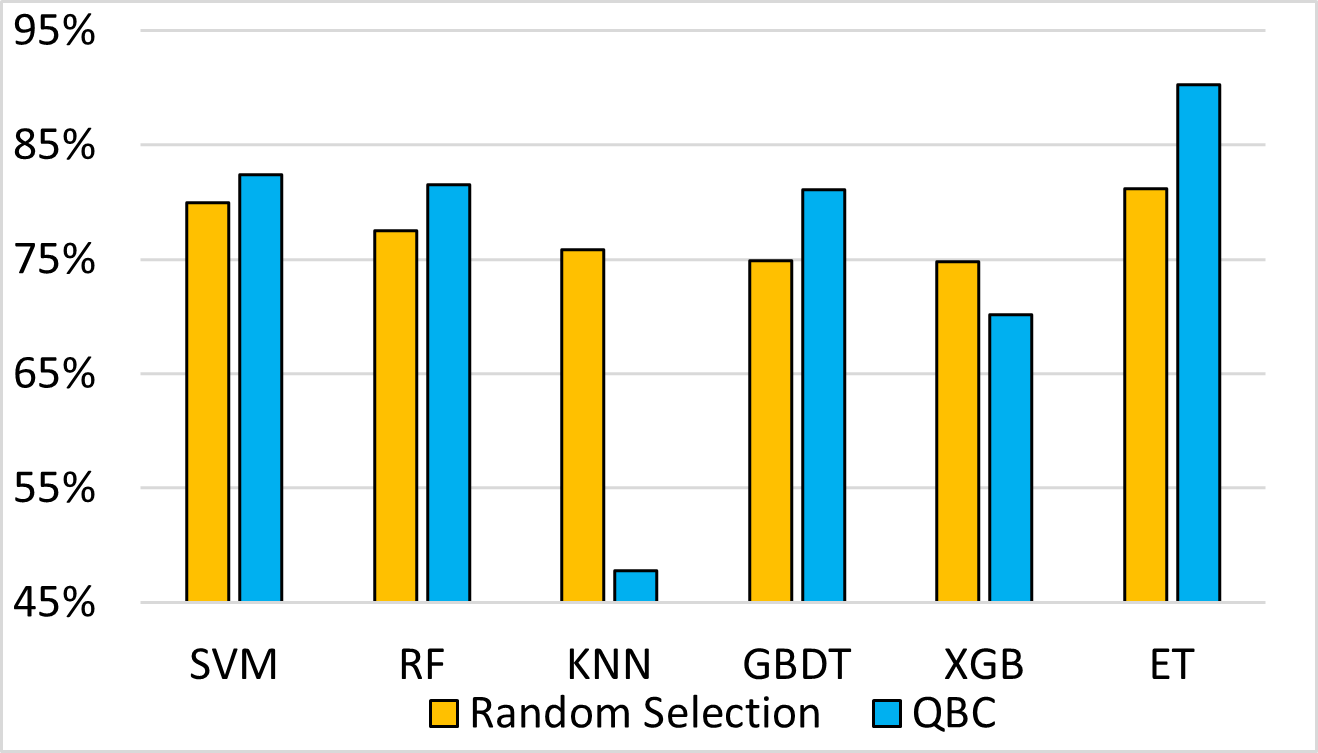}
%\caption{fig1}
\label{Accuracy_11}
\end{minipage}%
}%
\subfigure[SVHN (ResNet-18 is the teacher model)]{
\begin{minipage}[t]{0.33\linewidth}
\centering
\includegraphics[width=5.6cm]{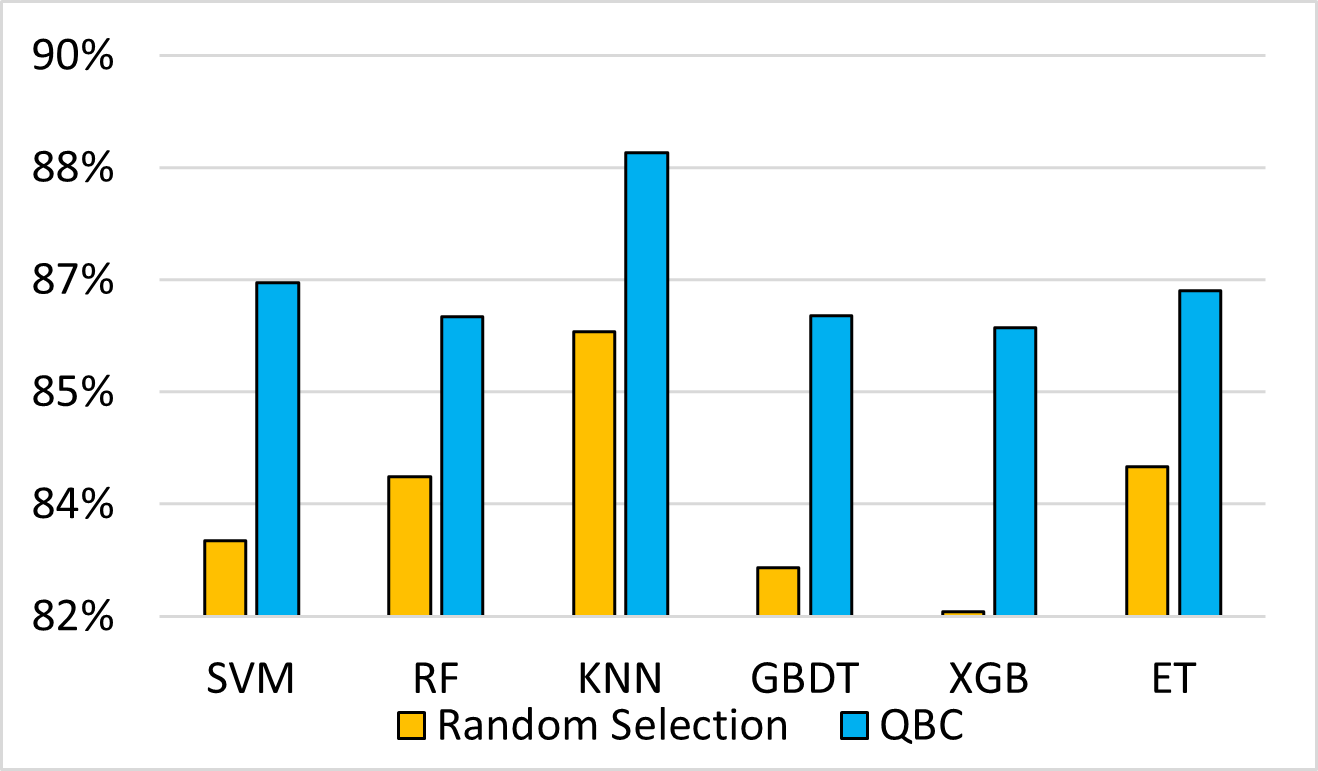}
\label{Accuracy_22}
%\caption{fig2}
\end{minipage}%
}%
\subfigure[CIFAR-10 (ResNet-18 is the teacher model)]{
\begin{minipage}[t]{0.33\linewidth}
\centering
\includegraphics[width=5.6cm]{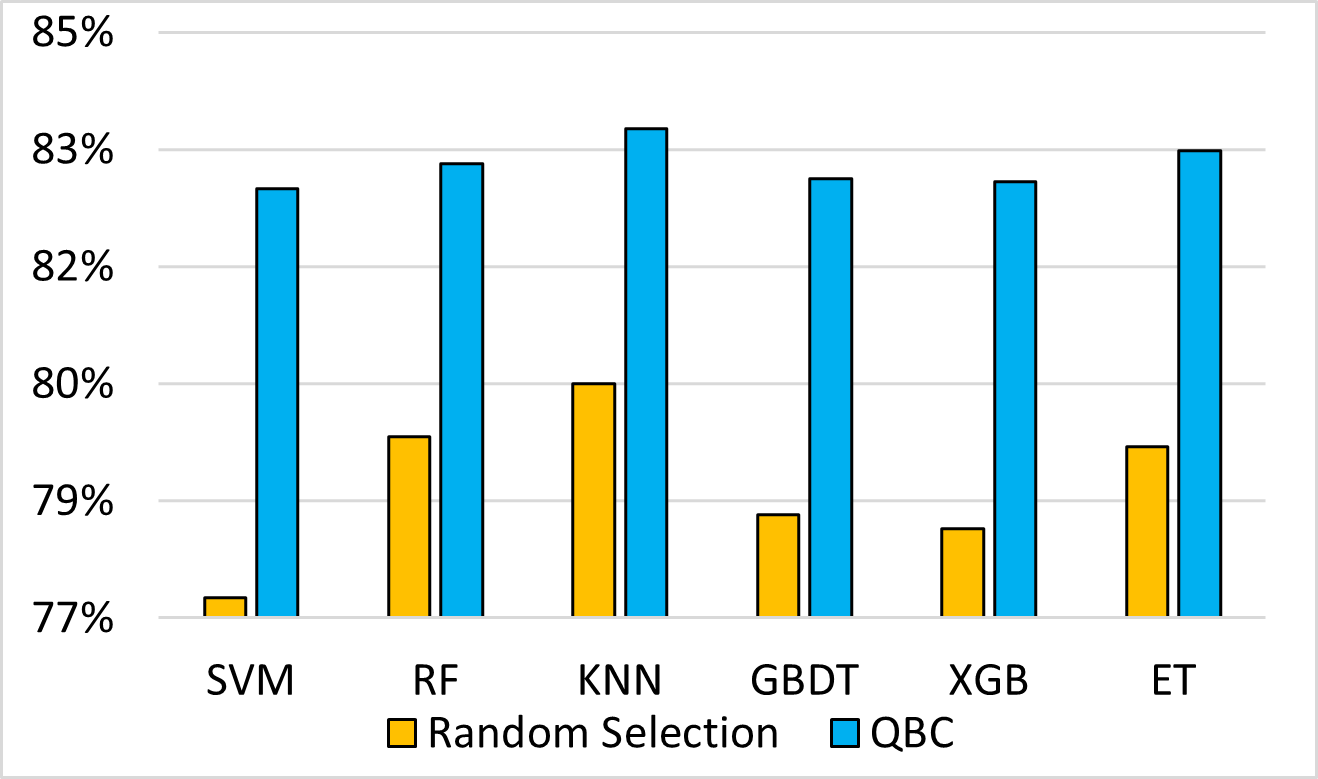}
\label{Accuracy_33}
%\caption{fig2}
\end{minipage}%
}%

\subfigure[MNIST (ResNet-34 is the teacher model)]{
\begin{minipage}[t]{0.33\linewidth}
\centering
\includegraphics[width=5.6cm]{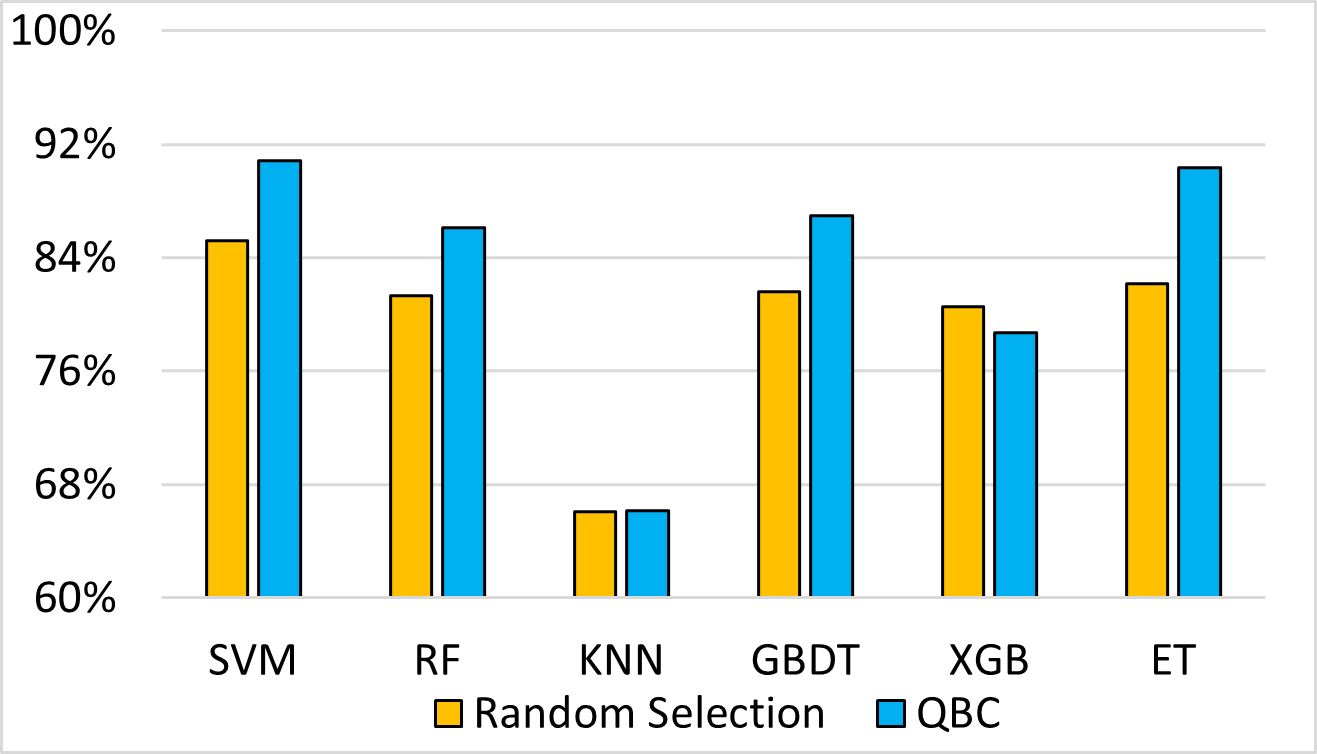}
\label{Accuracy_1}
%\caption{fig1}
\end{minipage}%
}%
\subfigure[SVHN (ResNet-34 is the teacher model)]{
\begin{minipage}[t]{0.33\linewidth}
\centering
\includegraphics[width=5.6cm]{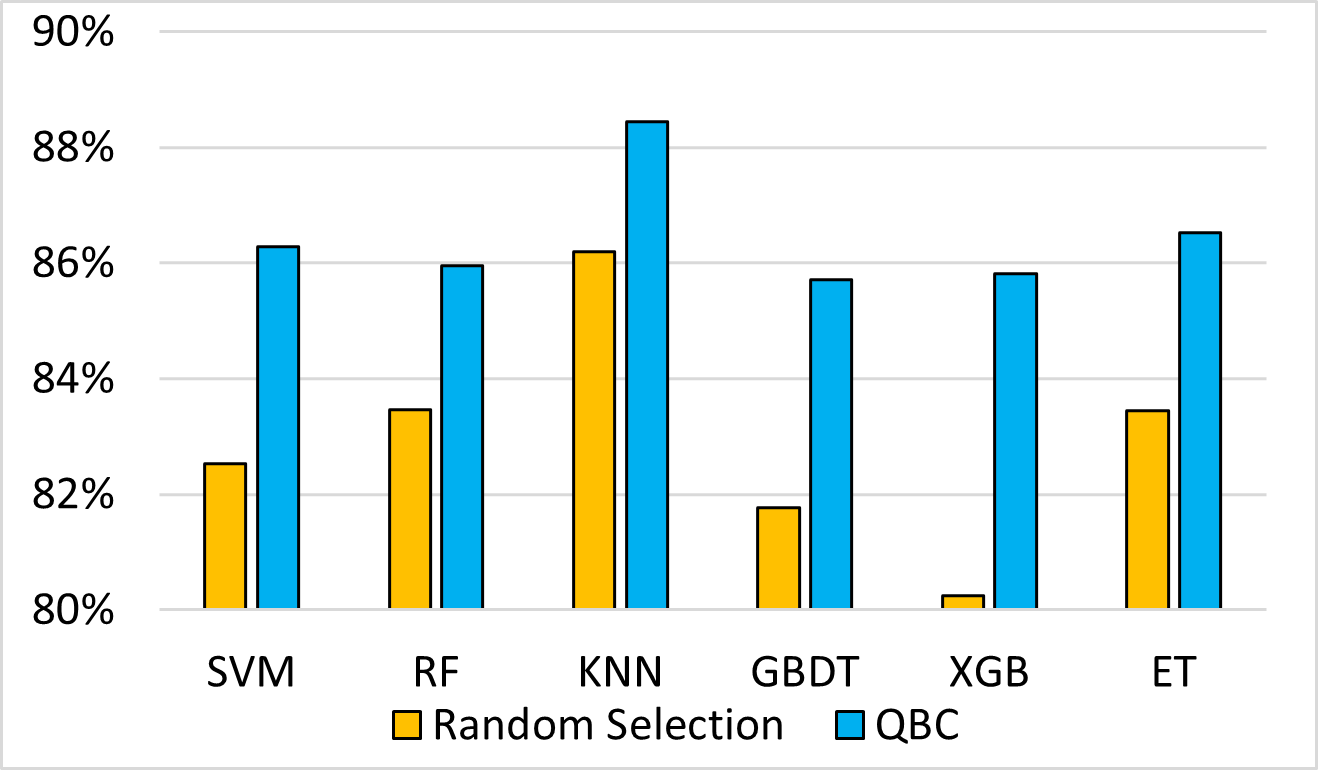}
\label{Accuracy_2}
%\caption{fig2}
\end{minipage}%
}%
\subfigure[CIFAR-10 (ResNet-34 is the teacher model)]{
\begin{minipage}[t]{0.33\linewidth}
\centering
\includegraphics[width=5.6cm]{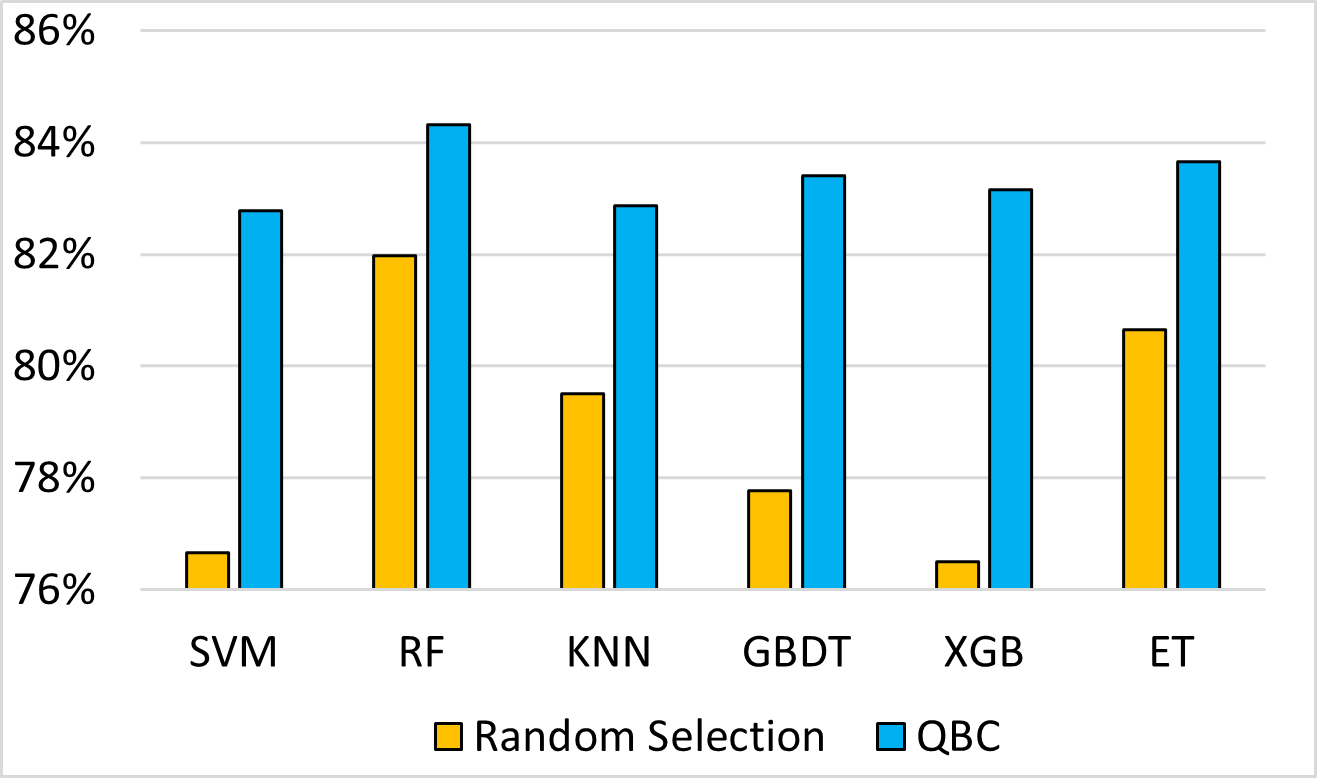}
\label{Accuracy_3}
%\caption{fig2}
\end{minipage}%
}%
\vspace{-9pt}
\caption{Accuracy of SL-based Offloading Strategy\vspace{-10pt}}
\label{Accuracy}
\end{figure*}

We evaluate the performance of SL-based offloading strategies after obtaining student models. According to Algorithm~\ref{weidaima2}, $M$ complex samples are first obtained using student models. Then, random selection and QBC are applied to choose normal data, respectively. Finally, various kinds of SL models are trained and tested based on Fig.~\ref{offloading2}. The accuracy of SL models is listed in Fig.~\ref{Accuracy}.

Fig.~\ref{Accuracy} shows that SL models trained by QBC algorithms outperform those trained by random selection. The results demonstrate that QBC algorithms can find those data that are more informative for SL models. Besides, it can be seen that the SL model which obtains the highest accuracy varies depending on the datasets and teacher models. For example, on MNIST, ET achieves the highest accuracy for ResNet-18 while SVM reaches the optimal performance for ResNet-34; on SVHN, KNN achieves the highest accuracy for both ResNet-18 and ResNet-34; on CIFAR-10, KNN and RF obtain the highest accuracy for ResNet-18 and ResNet-34, respectively. In addition to accuracy, the size of SL models should also be considered as it affects the execution time on the edge. Table~\ref{SLsize18} and~\ref{SLsize34} list the size of the SL models. Take CIFAR-10 as an example, although RF has the highest accuracy for ResNet-34, its size is larger than 10 MB, hence leading to high computation latency. In contrast, the sizes of other SL models except ET are no more than 500 KB but obtain similar accuracy. In a word, the tradeoff between accuracy and size should be considered when evaluating the SL models. 

In a real-world implementation, it is inconvenient to train and test numerous SL models and then to analyze which SL model is the optimal offloading strategy. Therefore, another way to select SL models is to choose the SL model that shows stable performances and has a small size no matter what the datasets are. From Fig.~\ref{Accuracy} and Table~\ref{SLsize18}~\ref{SLsize34}, it can be observed that the accuracy of SVM reaches the average level of all SL models, and the size of SVM is the smallest among the SL models in most cases. Thus, in practical applications, we can use SVM as a general offloading strategy. 
% \toprule
% \midrule
% \bottomrule 

\begin{table}[htbp]
\caption{Size (KB) of SL models (ResNet-18 is the teacher model)\vspace{-8pt}
}
\renewcommand\arraystretch{1}
\setlength{\tabcolsep}{1.36mm}{
\begin{tabular}{ccccccc}\toprule
Dataset & SVM        & RF             & KNN          & GBDT & XGB & ET    \\\midrule
MNIST   & 6 & 197            & 14           & 151  & 73  & 496   \\
SVHN    & 138        & 4417           & 425 & 172  & 317 & 16149 \\
CIFAR-10   & 467        & 12781 & 2009      & 174  & 404 & 44390 \\\bottomrule 
\end{tabular}}
\label{SLsize18}
\end{table}

\begin{table}[htbp]
\caption{Size (KB) of SL models (ResNet-18 is the teacher model)\vspace{-8pt}
}
\renewcommand\arraystretch{1}
\setlength{\tabcolsep}{1.5mm}{
\begin{tabular}{ccccccc}\toprule
Dataset & SVM        & \multicolumn{1}{c}{RF} & KNN          & GBDT & XGB & ET    \\\midrule
MNIST   & 7 & 213                    & 12           & 149  & 81  & 490   \\
SVHN    & 122        & 3800                   & 507 & 171  & 307 & 14050 \\
CIFAR-10   & 116        & 3738          & 425          & 174  & 294 & 13375\\\bottomrule 
\end{tabular}}
\label{SLsize34}
\end{table}

Furthermore, the sizes of SL models are listed in Table~\ref{size1} and~\ref{size2}. It can be seen that the total sizes of SL models and student models are still much smaller than the student models generated in N2N learning, which implies that our method requires fewer computation resources.
% （为什么有的数据要加粗？QBC比随机选择似乎只好一点点？有没有更合适的性能指标表达data offload判断complex sample的准确性？）

% \toprule
% \midrule
% \bottomrule 

\begin{table}[htbp]
\caption{Size of offloading strategies and student models (ResNet-18 is the teacher model)}
\renewcommand\arraystretch{1}
\setlength{\tabcolsep}{2mm}{
\begin{tabular}{ccccc}\toprule
                          & \multicolumn{3}{c}{Proposed Method}           &                                \\
\multirow{-2}{*}{Dataset} & Student              & SL & Total & \multirow{-2}{*}{N2N} \\\midrule
MNIST                     & 407 KB   & 6 KB        & 413 KB     & \diagbox[width=3.5em]{}{}               \\
SVHN                      & 491 KB                          & 138 KB        & 629 KB     & 2239 KB                             \\
CIFAR-10                  & 675 KB                          & 467 KB       & 1142 KB    & 3963 KB                \\\bottomrule             
\end{tabular}}
\label{size1}
\end{table}

\begin{table}[htbp]

\caption{Size of offloading strategies and student models (ResNet-34 is the teacher model)
}
\renewcommand\arraystretch{1}
\setlength{\tabcolsep}{2mm}{
\begin{tabular}{ccccc}\toprule
                          & \multicolumn{3}{c}{Proposed Method}           &                                \\
\multirow{-2}{*}{Dataset} & Student              & SL & Total & \multirow{-2}{*}{N2N} \\\midrule
MNIST                     & 818 KB   & 7 KB        & 825 KB     & \diagbox[width=3.5em]{}{}               \\
SVHN                      & 785 KB   & 122 KB        & 907 KB     & \diagbox[width=3.5em]{}{}                            \\
CIFAR-10                  & 1769 KB   & 116 KB       & 1885 KB    & 8156 KB                \\\bottomrule             
\end{tabular}}
\label{size2}
\end{table}

\begin{table*}[]
\caption{Accuracy and runtime under different edge-cloud systems on real hardware (ResNet-18 is the teacher model)
}
\centering
\renewcommand\arraystretch{1}
\setlength{\tabcolsep}{5mm}{
\begin{tabular}{ccccccc}\toprule
\multirow{2}{*}{Dataset} & \multicolumn{2}{c}{Only edge} & \multicolumn{2}{c}{Only cloud} & \multicolumn{2}{c}{Cooperation (Proposed)} \\ 
                         & Runtime    & Accuracy   & Runtime    & Accuracy    & Runtime         & Accuracy         \\ \midrule
MNIST                    & 48.6 s                & 99.49\%          & 255.5 s                & 99.34\%          & 54.2 s                    & 99.74\%              \\ 
SVHN                     & 196.1 s                & 93.19\%          & 1407.5 s                & 95.1\%           & 313.7 s                     & 98.1\%               \\ 
CIFAR-10                 & 174.2 s                & 84.12\%          & 331.9 s                & 88.73\%          & 182.2 s                     & 93.45\%               \\ \bottomrule   
\end{tabular}}
\label{hardware18}
\end{table*}

\begin{table*}[]
\caption{Accuracy and runtime under different edge-cloud systems on real hardware (ResNet-34 is the teacher model)
}
\centering
\renewcommand\arraystretch{1}
\setlength{\tabcolsep}{5mm}{
\begin{tabular}{ccccccc}\toprule
\multirow{2}{*}{Dataset} & \multicolumn{2}{c}{Only Edge} & \multicolumn{2}{c}{Only Cloud} & \multicolumn{2}{c}{Cooperation (Proposed)} \\ 
                         & Runtime    & Accuracy   & Runtime    & Accuracy    & Runtime         & Accuracy         \\ \midrule
MNIST                    & 89.3 s                & 99.44\%          & 397.7 s                & 99.51\%          & 99.1 s                     & 99.76\%              \\ 
SVHN                     & 485.7 s                & 92.91\%          & 1211.8 s                & 95.08\%           & 675.9 s                     & 97.81\%               \\ 
CIFAR-10                 & 261.9 s                & 86.20\%          & 371.9 s               & 92.32\%          & 280.3 s                     & 94.6\%               \\ \bottomrule   
\end{tabular}}
\label{hardware34}
\end{table*}
% 在table多加一列SL 

\subsection{Edge-cloud framework performance on real hardware}

We implement the proposed framework on real hardware on three datasets. Two criteria, runtime and accuracy, are used for performance evaluation. Firstly, we only execute the student model on the edge and overlook the cloud. Secondly, all the data are offloaded to the cloud center and the teacher models are used to carry out DNN inference. Thirdly, we apply the proposed edge-cloud cooperation framework. The experiment results are listed in the Table~\ref{hardware18} and \ref{hardware34}. For accuracy, the proposed edge-cloud cooperation framework obtains the highest accuracy as some complex data are offloaded to the cloud server according to the SL-based offloading strategy. As mentioned in Section~\ref{intro}, the student models can provide correct inference results for some samples while the teacher models can't~\cite{Anubhav18}. Thus, the accuracy of the whole system is higher than its corresponding teacher model. In terms of runtime, offloading all the data to the cloud causes a large delay due to data transmission, while the only edge-only strategy has the shortest runtime at the expense of accuracy. To sum up, the extra delay introduced by edge-cloud cooperation is quite small (less than 20\% in most cases) compared with the edge-only strategy, while the accuracy is considerably improved; compared with the cloud-only strategy, the inference latency is greatly reduced with even higher accuracy.

%More details are provided in Section~\ref{sl}.

% \section{Discussion}
% In our experiments, the SL models for offloading strategies are determined based on the .

\section{Conclusion}~\label{section7}
In this paper, we present an edge-cloud cooperation framework to achieve high accuracy but low inference time for DNN inference. The proposed framework contains two key technologies: RL-based DNN compression and SL-based offloading strategies. The RL-based DNN compression generates a student model from a teacher model. The student model is deployed on resource-limited edge devices and can handle most of the data. As the accuracy of the student model is lower than that of the teacher model, the SL-based offloading strategy is proposed to counterbalance this deterioration. The SL-based offloading strategy identifies the complex data for the student model and sends them to the cloud. Extensive experiments have demonstrated that our framework achieves the highest accuracy compared with both Edge-Only and Cloud-Only strategies, while greatly reducing the inference latency for data offloading.

\section*{Acknowledgement}
This research is supported by the National Research Foundation, Singapore under its Strategic Capability Research Centres Funding Initiative. Any opinions, findings and conclusions or recommendations expressed in this material are those of the author(s) and do not reflect the views of National Research Foundation, Singapore. This work was also supported in part by the NTU-Wallenberg AI, Autonomous Systems and Software Program (WASP) Joint Project.
\ifCLASSOPTIONcaptionsoff
  \newpage
\fi



%\bibliographystyle{IEEEtran}
\bibliographystyle{IEEEtran}
\bibliography{refer}

\begin{thebibliography}{10}
\providecommand{\url}[1]{#1}
\csname url@samestyle\endcsname
\providecommand{\newblock}{\relax}
\providecommand{\bibinfo}[2]{#2}
\providecommand{\BIBentrySTDinterwordspacing}{\spaceskip=0pt\relax}
\providecommand{\BIBentryALTinterwordstretchfactor}{4}
\providecommand{\BIBentryALTinterwordspacing}{\spaceskip=\fontdimen2\font plus
\BIBentryALTinterwordstretchfactor\fontdimen3\font minus
  \fontdimen4\font\relax}
\providecommand{\BIBforeignlanguage}[2]{{%
\expandafter\ifx\csname l@#1\endcsname\relax
\typeout{** WARNING: IEEEtran.bst: No hyphenation pattern has been}%
\typeout{** loaded for the language `#1'. Using the pattern for}%
\typeout{** the default language instead.}%
\else
\language=\csname l@#1\endcsname
\fi
#2}}
\providecommand{\BIBdecl}{\relax}
\BIBdecl

\bibitem{Anubhav18}
{Ashok, Anubhav and Rhinehart, Nicholas and Beainy, Fares and Kitani, Kris},
  ``N2n learning: Network to network compression via policy gradient
  reinforcement learning,'' in \emph{Proceedings of the 6th International
  Conference on Learning Representations}, 2018.

\bibitem{Kang17}
{Yiping Kang, Johann Hauswald, Cao Gao, Austin Rovinski, Trevor Mudge, Jason
  Mars, and Lingjia Tang}, ``Neurosurgeon: Collaborative intelligence between
  the cloud and mobile edge,'' in \emph{Proceedings of the 25th International
  Conference on Architectural Support for Programming Languages and Operating
  Systems}, 2017, p. 615–629.

\bibitem{En18}
{En Li, Zhi Zhou, and Xu Chen}, ``Edge intelligence: On-demand deep learning
  model co-inference with device-edge synergy,'' in \emph{Proceedings of the
  2018 Workshop on Mobile Edge Communications}, 2018, p. 31–36.

\bibitem{Surat17}
{Surat Teerapittayanon and Bradley McDanel and H. T. Kung}, ``Branchynet: Fast
  inference via early exiting from deep neural networks,'' in \emph{arXiv},
  2017.

\bibitem{Teerapittayanon17}
{S. {Teerapittayanon} and B. {McDanel} and H. T. {Kung}}, ``Distributed deep
  neural networks over the cloud, the edge and end devices,'' in \emph{2017
  IEEE 37th International Conference on Distributed Computing Systems (ICDCS)},
  2017, pp. 328--339.

\bibitem{Zhang20}
{Zhang, Shigeng and Li, Yinggang and Liu, Xuan and Guo, Song and Wang, Weiping
  and Wang, Jianxin and Ding, Bo and Wu, Di}, ``Towards real-time cooperative
  deep inference over the cloud and edge end devices,'' \emph{Proc. ACM
  Interact. Mob. Wearable Ubiquitous Technol.}, vol.~4, no.~2, Jun. 2020.

\bibitem{Song16}
S.~Han, H.~Mao, and W.~Dally, ``Deep compression: Compressing deep neural
  networks with pruning, trained quantization and huffman coding,'' in
  \emph{Proceedings of the 4th International Conference on Learning
  Representations}, 2016.

\bibitem{Kim16}
{ Kim, Yong Deok and Park, Eunhyeok and Yoo, Sungjoo and Choi, Taelim and Yang,
  Lu and Shin, Dongjun }, ``Compression of deep convolutional neural networks
  for fast and low power mobile applications,'' in \emph{Proceedings of the 4th
  International Conference on Learning Representations}, 2016.

\bibitem{Yang19}
{Yang, Haichuan and Zhu, Yuhao and Liu, Ji}, ``End-to-end learning of
  energy-constrained deep neural networks,'' in \emph{Proceedings of the 7th
  International Conference on Learning Representations}, 2019.

\bibitem{Liang19}
L.~Huang, S.~Bi, and Y.-J. Zhang, ``Deep reinforcement learning for online
  computation offloading in wireless powered mobile-edge computing networks,''
  vol.~19, pp. 2581--2593, 2020.

\bibitem{Fariba20}
F.~Farahbakhsh, A.~Shahidinejad, and M.~Ghobaei-Arani, ``Multiuser
  context-aware computation offloading in mobile edge computing based on
  bayesian learning automata,'' \emph{Transactions on Emerging
  Telecommunications Technologies}, pp. 1--26, 2020.

\bibitem{Jiang20}
{F. {Jiang} and K. {Wang} and L. {Dong} and C. {Pan} and K. {Yang}}, ``Stacked
  autoencoder-based deep reinforcement learning for online resource scheduling
  in large-scale mec networks,'' \emph{IEEE Internet of Things Journal},
  vol.~7, no.~10, pp. 9278--9290, 2020.

\bibitem{Williams92}
R.~J. Williams, ``Simple statistical gradient-following algorithms for
  connectionist reinforcement learning,'' \emph{Mach. Learn.}, vol.~8, no.
  3–4, p. 229–256, May 1992.

\bibitem{Seung92}
{Seung, H. S. and Opper, M. and Sompolinsky, H.}, ``Query by committee,'' in
  \emph{Proceedings of the Fifth Annual Workshop on Computational Learning
  Theory}, 1992, p. 287–294.

\bibitem{Tung19}
F.~{Tung} and G.~{Mori}, ``Similarity-preserving knowledge distillation,'' in
  \emph{2019 IEEE/CVF International Conference on Computer Vision (ICCV)},
  2019, pp. 1365--1374.

\end{thebibliography}


\end{document}